\documentclass{article}

\usepackage[final]{neurips_2024}

\usepackage[utf8]{inputenc} 
\usepackage[T1]{fontenc}    
\usepackage{hyperref}       
\usepackage{url}            
\usepackage{booktabs}       
\usepackage{amsfonts}       
\usepackage{amsmath}
\usepackage{nicefrac}       
\usepackage{microtype}      
\usepackage{xcolor}         
\usepackage{natbib}
\usepackage{graphicx}
\usepackage[inline,shortlabels]{enumitem}
\usepackage{fontawesome}

\newcommand{\dat}{\textbf{d}}
\newcommand{\parvec}{{\theta}}

\usepackage[normalem]{ulem}

\title{Hybrid Summary Statistics}

\author{%
T. Lucas Makinen$^{1*}$ \quad Ce Sui$^{2}$\thanks{Equal contribution} \quad Benjamin D. Wandelt$^{3,4}$ \\ 
\textbf{Natalia Porqueres}$^5$ \quad \textbf{Alan Heavens}$^1$ \\
$^1$Imperial College London \quad $^2$Tsinghua University \quad $^3$Sorbonne Universit\'e \\ 
$^4$Center for Computational Astrophysics, Flatiron Institute\\
$^5$Oxford University \\
\texttt{l.makinen21@imperial.ac.uk}\\
\texttt{suic20@mails.tsinghua.edu.cn}\\
}

\begin{document}

\maketitle

\begin{abstract}
    We present a way to capture high-information posteriors from training sets that are sparsely sampled over the parameter space for robust simulation-based inference. In physical inference problems, we can often apply domain knowledge to define traditional summary statistics to capture some of the information in a dataset. We show that augmenting these statistics with neural network outputs to maximise the mutual information improves information extraction compared to neural summaries alone or their concatenation to existing summaries and makes inference robust in settings with low training data. We introduce 1) two loss formalisms to achieve this and 2) apply the technique to two different cosmological datasets to extract non-Gaussian parameter information.
\end{abstract}

\begin{figure}[htp!]
    \centering
    \includegraphics[width=0.96\linewidth]{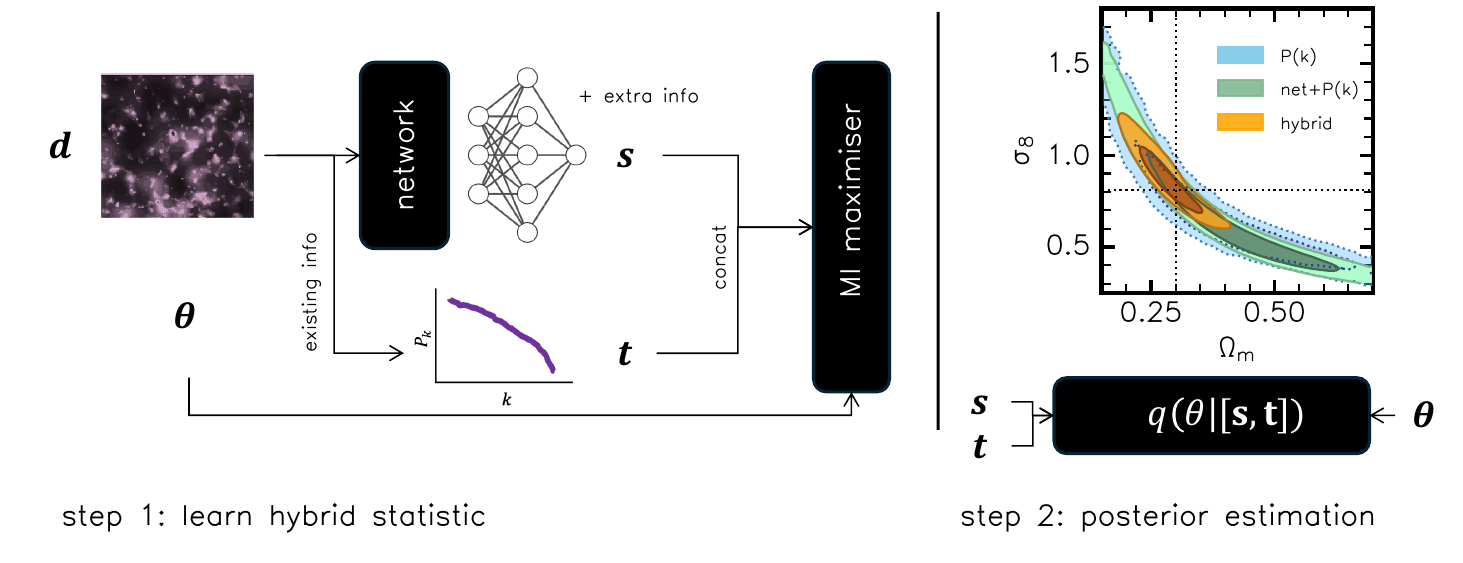}
    \caption{Schematic for learning a hybrid summary statistic $\textbf{s}$ with a choice of mutual information maximiser and existing static \textbf{t}. Black boxes denote neural functions to be learned.}
    \label{fig:schematic}
\end{figure}

\section{Introduction}\label{sec:method}
Implicit (simulation-based) inference makes solving otherwise intractable inverse problems possible by employing neural compressors which can flexibly map big data vectors into informative summaries \citep{Cranmer30055, hoffmann2023minimising}. These mappings can be made optimal \citep{Charnock_IMNN, makinen2021, makinen2024hybridsummarystatisticsneural, lanzieri2024optimalneuralsummarisationfullfield, Jeffrey_massmap_vmim}, but often require large numbers of simulations to achieve convergence. 

For many physics problems, such as large N-body and hydrodynamical solvers, forward simulations are expensive to generate, so networks tasked to compress these data for parameter inference must learn informative features using sparsely-sampled training sets, especially when large numbers of parameters are varied. {}{Introducing informative priors in data feature \citep[e.g.]{battaglia2018relational, ivanov2024fullshapeanalysissimulationbasedpriors} or likelihood \citep{modi2023hybridsbiilearned} space can simplify the task of a neural network to learn an objective.}

\noindent \textbf{Main Contributions.} We present a way to obtain highly informative summaries over parameter space in low-training data settings that \textit{boost} information extraction from data beyond existing statistics and the mere concatenation of neural summaries learned separately and traditional summaries. These ``hybrid" statistics are neural summaries that are learned to maximise the mutual information (MI) beyond an existing summary of the data and the parameters of interest.

\noindent \textbf{Summary of Results.} We present two equivalent objectives that maximise MI information between a new, neural summary, and an existing summarisation of the data over parameter space. We apply this technique to two different cosmological problems where information can be lost to existing summary functions--21cm Epoch of Reionisation (21cm) and weak gravitational lensing (WL) parameter inference. We show that hybridised summaries are far more robust in settings with low available training simulations, indicating improved network optimisation over parameter space.

\section{Formalism}
 In inference problems, we aim to infer quantities of interest $\parvec$ from data $\dat$. In the physical sciences we often have domain knowledge which allows us to design a summary statistic of the data $\textbf{t}(\dat)$ that capture some, but not all, of the information in a dataset. To enhance the information extraction, we aim to learn additional summaries $\textbf{s}(\dat)$ that are complementary to $\textbf{t}$. We can formalise the extra information beyond $\textbf{t}$ captured by $\textbf{s}$ as the conditional mutual information $I(\textbf{s}; \parvec | \textbf{t})$, which measures how much the uncertainty about $\parvec$ is reduced by knowing $\textbf{s}$ given a static $\textbf{t}$. Evaluating the conditional mutual information can be difficult. However, using the chain rule of MI we can write
\begin{equation}
    I(\textbf{s} ; \parvec | \textbf{t}) = I([\textbf{s}, \textbf{t}]; \parvec) - I(\textbf{t}; {\parvec})
\end{equation}
where $I(\textbf{t}; \parvec)$ is a constant and $[a,b]$ denotes concatenation. We can then maximise the additional information captured by $\textbf{s}$ by maximising $I([\textbf{t}, \textbf{s}]; \parvec)$. There are various ways to maximize MI; here, we focus on two specific objectives, detailed in Appendix~\ref{ap:MI maximization}. The first, referred to as the Posterior Entropy (EPE) objective, is given by:
\begin{equation}
\label{eq:epe_loss}
     \min_{\textbf{s},q} \mathcal{L}
    = -\mathbb{E}_{p(\theta, \dat)}\Big[\log q\left(\theta \big| [\textbf{s}(\dat), \textbf{t}(\dat)]\right)\Big],
\end{equation}
where $\textbf{s}(\textbf{d})$ is a neural network and $q$ is a neural density estimator. \citet{hoffmann2023minimising} demonstrate that this loss unifies many information-theoretic losses into a stable objective. This mutual information objective can also be parameterised as a cross-entropy classification problem by employing the Jensen-Shannon divergence \citep{chen2021neuralapproximatesufficientstatistics, infomax,fGAN}:
\begin{equation}
\label{eq:ce_loss}
    \min_{\textbf{s}, c} \mathcal{L}(\textbf{s}, c) = \mathbb{E}_{p(\theta, \dat)} \left[\mathrm{sp}(-c(\theta, [\textbf{s}(\dat), \textbf{t}(\dat)]))\right] + \mathbb{E}_{p(\theta)p(\dat)} \left[\mathrm{sp}(c(\theta, [\textbf{s}(\dat), \textbf{t}(\dat)]))\right],
\end{equation}
where $s$ is the summarizer, $c$ is a classifier tasked with distinguishing between data from $p(\theta, x)$ and $p(\theta)p(x)$ and $\mathrm{sp}(z) = \log(1 + e^z)$ is the softplus function. We refer to this as the Cross Entropy (CE) loss.

To test the information capture in learned summaries we employ a two-step process, illustrated in Fig. \ref{fig:schematic}. We first optimise a neural compression (embedding network) to a few additional numbers conditional on the specified (not learned) existing summary $\textbf{t}$, using either EPE or CE losses (denoted MI maximiser in Fig. \ref{fig:schematic}). For the EPE loss we train a simple mixture density network (MDN) with two small hidden layers to approximate $q(\theta | [\textbf{s}, \textbf{t}])$. For the CE loss we train a classifier fully-connected network with hidden sizes $[128,64,64]$ to one output. We then take the static learned and existing summaries and parameterise a separate posterior estimator using a masked autoregressive flow (MAF) to minimise $p(\theta | [\textbf{s}, \textbf{t}])$ from the  {\tt LtU-ILI} package\footnote{https://github.com/maho3/ltu-ili} \citep{ho2024ltuili}. We feed comparison summaries (power spectrum and competing network schemes) into the same MAF architecture to obtain a consistent comparison of information capture. We detail exact configurations in Appendix \ref{ap:sim_net_details}.

\section{Experiments}
\noindent \textbf{21cm Parameter Inference \& Loss Comparison.}
The 21~cm signal is non-Gaussian due to reionization patchiness. Therefore, the power spectrum alone cannot fully capture the information contained in images of the 21 cm signal. While many previous studies have focused on designing new summaries, we show that adding just a few supplementary features to the power spectrum can significantly enhance the extracted information content. 

The reionization parameters we vary are \begin{enumerate*}
 \item[(1)] $\zeta$, the ionizing efficiency. It primarily determines the timing of the EoR, with higher values leading to earlier ionization of the Universe. We vary $\zeta$ as $10 \le \zeta \le 250 $, and

\item[(2)] $T_ { \mathrm { vir } }$, the minimum virial temperature of halos that host ionizing sources. $T_{\mathrm{vir}}$ controls the timing of astrophysical epochs and influences the scales of heated and ionized regions. We vary this parameter as $ 4 \le \log _ { 10 } \left( T_{ \mathrm { vir } } / \mathrm { K } \right) \le 6 $.
 \end{enumerate*} More details can be found in Appendix \ref{ap:sim_net_details}.

\begin{figure}
  \centering
  \includegraphics[width=0.86\linewidth]{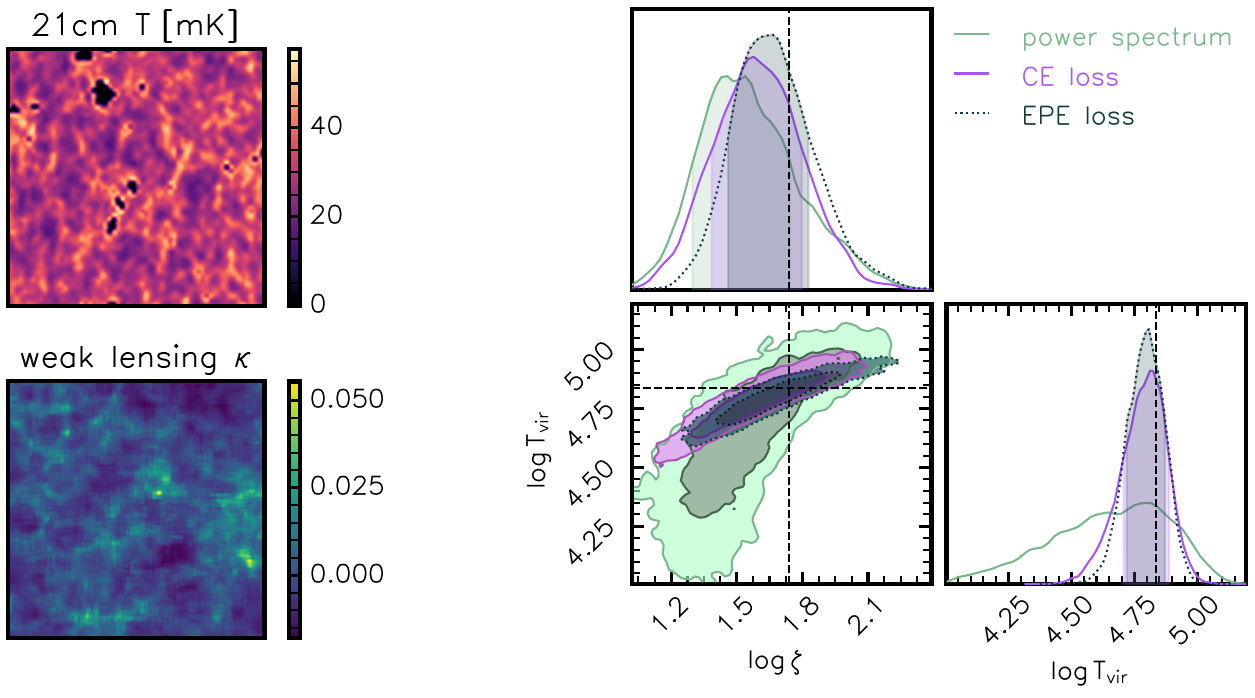}
  \caption{Both 21cm and weak lensing data exhibit non-Gaussian features (upper and lower left panels). Both EPE and CE loss formalisms result in consistent, tighter posteriors than power spectrum alone  indicating information extraction from non-Gaussian features in the 21cm data (right). The black dashed line represents the true parameter values.
  }
  \label{fig:21cm_results}
\end{figure}

 In this initial experiment, we compare tree types of summaries, assuming a sufficiently large training set (10,000 samples): \begin{enumerate*}

 \item[1)] Power spectrum only (11 $k$-bins).

\item[2)] Hybrid method: Power spectrum + two learned supplementary features (EPE Loss).

\item[3)]Hybrid method: Power spectrum + two learned supplementary features (CE Loss).


One of the resulting inferences is shown in Figure~\ref{fig:21cm_results}. The results show that both approaches capture non-Gaussian information and improve inference performance, yielding {}{consistent posteriors}. {}{The slight difference in posterior recovery is likely due to differences between classifier and MDN network architectures for each loss, but each yielded similar convergence times in training.} This suggests we can learn two additional parameters instead of new summaries, maintaining power spectrum interpretability while achieving near-optimal inference. The agreement also confirms both loss functions are effective, converging to the same results.

 \end{enumerate*}

\begin{figure}[htp!]
    \centering
    \includegraphics[width=0.9\linewidth]{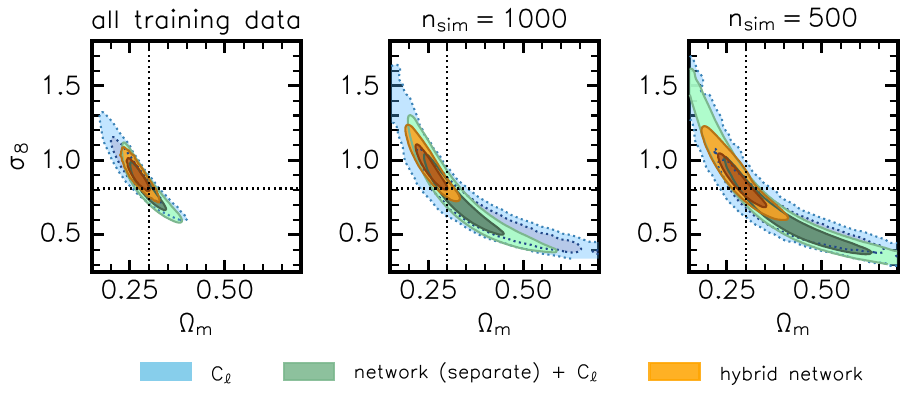}
    \caption{Hybrid summaries are able to capture more non-Gaussian parameter information in settings with both high (left) and low (right) numbers of training simulations than {}{summaries from a} larger network trained with the same loss separately from and then concatenated to the $C_\ell$ statistic {}{for the final inference in step 2}. When reducing the total available simulation volume all inferences suffer, but the hybrid statistics are most robust to this change, indicating that the MI objective improves capture of non-power spectrum features.}
    \label{fig:wl-ablation}
\end{figure}

\noindent \textbf{Tomographic Weak Lensing Inference \& Ablation Study.} Weak gravitational lensing (WL) alters the trajectories of photons as they pass through massive structures of visible and dark matter. This observable is sensitive to $(\Omega_m, S_8$), parameters that control the universe's matter content and dark matter clustering, respectively. Here we test hybrid summaries on noisy tomographic WL convergence image data of shape $(128,128,4)$ presented in \cite{makinen2024hybridsummarystatisticsneural} varied over a wide uniform prior. As an existing summary, we repeat \citet{makinen2024hybridsummarystatisticsneural}'s procedure and histogram all auto- and cross-power spectra for each redshift bin into a vector of 60 numbers for each simulation. The simulations are also subject to additive shape noise, which we add to the noise-free simulations on-the-fly during network training (details in Appendix \ref{ap:sim_net_details}). 

We perform an ablation study to demonstrate the effectiveness of the hybrid statistics over neural-only methods and display results in Fig. \ref{fig:wl-ablation}. The embedding network in the hybrid setting is the lightweight CNN with symmetric kernels adapted from \cite{makinen2024hybridsummarystatisticsneural} to output 3 additional numbers alongside the $C_\ell$s. For comparison, we train a larger, more expressive CNN embedding network under the same EPE loss \textit{without} access to the $C_\ell$ vector to output 3 numbers, and then for the density estimation (step 2) concatenate its outputs to the $C_\ell$s (green contours; network (separate) + $C_\ell$).  Step 2 probes the information content captured in the static network and $C_\ell$ summaries. We train the networks and density estimators {}{from scratch} first using all $5000$ simulations available (split into 70\% train and 30\% validation sets). We then reduce the total number of available simulations to $1000$ and further to $500$ and re-learn the embeddings and posteriors from scratch. When the simulation budget is reduced, all inferences suffer, but the hybrid statistic formalism encourages the smaller network to find non-Gaussian features in the dataset that are more robust to this change. We also note that the more expressive, separately-trained CNN inference degrades almost to the level of the $C_\ell$ contour in the lowest-data setting, even when concatenated to the $C_\ell$ vector in the density estimation step.

\section{Conclusions \& Outlook}
We detailed a method to learn compressed summary statistics that explicitly complement an existing summary of the data through mutual information maximisation. We show that these techniques can capture non-Gaussian information in two cosmological applications using two different loss criteria to significantly improve parameter information capture.

We additionally demonstrate that these hybridised summaries improve information capture when the training simulation budget is limited. This suggests that requiring a network to find patterns in the data that are explicitly complementary to a provided summary ``tells it where to look" and improves the compression optimisation in smaller datasets over wide parameter space.

We note that MI can also be used as a static metric to quantify the information content in arbitrary summaries as in \citet{Ce_2023}. This technique could be extended towards exhaustive information studies to measure how much more information might be unlocked with multiple traditional or neural statistics.

\section{Data Availability}
All code, tutorials, and relevant simulations can be found at \url{https://github.com/tlmakinen/hybridStats} \faGithub.

\section{Acknowledgements}
We thank Tao Jing, Justin Alsing, Tom Charnock, Niall Jeffrey, and David Spergel for conversations that inspired this work. This work was supported by the Simons Collaboration on “Learning the Universe”. TLM acknowledges the support of the Imperial College London President's Scholarship and G-Research for conference travel costs. CS is supported by the National SKA Program of China (grant No. 2020SKA0110401) and NSFC (grant No. 11821303).

\bibliography{mybib}
\bibliographystyle{aasjournal}

\appendix
\section{Mutual information maximization}
\label{ap:MI maximization}
Conditional mutual information (MI) is defined as
\begin{equation}
    I(\textbf{s}; \parvec | \textbf{t}) =\iiint p(\parvec,\textbf{s},\textbf{t}) \log\frac{p(\parvec|\textbf{s},\textbf{t})}{p(\parvec|\textbf{t})} d\textbf{s} d\textbf{t} d\parvec,
\end{equation}
which measures how much the uncertainty about $\parvec$ is reduced by knowing $\textbf{s}$ given $\textbf{t}$. Using the chain rule of MI we can write
\begin{equation}
    I(\textbf{s} ; \parvec | \textbf{t}) = I([\textbf{s}, \textbf{t}]; \parvec) - I(\textbf{t}; {\parvec})
\end{equation}
where $I(\textbf{t}; \parvec)$ is a constant and $[a,b]$ denotes concatenation. We can then maximise the additional information captured by $\textbf{s}$ by maximising $I([\textbf{t}, \textbf{s}]; \parvec)$.
Mutual information is defined as
\begin{equation}
    I(\parvec ; \textbf{z}) = D_{\rm KL}(p(\theta, \textbf{z}) \| p(\theta)p(\textbf{z})) = \mathbb{E}_{p(\theta, \textbf{z})} \left[ \log \frac{p(\theta | \textbf{z})}{p(\theta)} \right].
\end{equation}
where we package $\textbf{z}=[\textbf{s}, \textbf{t}]$. Most of the time, the true posterior $p(\theta | \textbf{z})$ is unknown. To address this, we approximate the posterior with a neural density estimator $q(\theta | \textbf{z})$, which provides a tractable lower bound on the MI \citep{ba-bound,VBofMI}:
\begin{equation}
    I(\theta; \textbf{z}) = \mathbb{E}_{p(\theta, \textbf{z})} \left[\log \frac{q(\theta | \textbf{z})}{p(\theta)}\right] + \mathbb{E}_{p(\textbf{z})} \left[D_{\mathrm{KL}}(p(\theta | \textbf{z}) \| q(\theta | \textbf{z}))\right] 
\geq \mathbb{E}_{p(\theta, \textbf{z})}[\log q(\theta | \textbf{z})] + h(\theta).
\end{equation}
This leads to a calculable loss function for maximizing mutual information, with $h(\theta)$ fixed, leading to the objective in Equation~\ref{eq:epe_loss}. 

Additionally, since the exact value of mutual information is not required, we can use alternative divergences that may offer better robustness and efficiency. By using the Jensen-Shannon divergence and the variational representation from \citet{fGAN}, we can derive a lower bound that leads to a cross-entropy loss (Equation~\ref{eq:ce_loss}), frequently used in representation learning \citep{chen2021neuralapproximatesufficientstatistics, infomax}.

\section{Experimental Details}\label{ap:sim_net_details}
\subsection{21cm Simulations}
The 21 cm signals are simulated using the publicly available code {\tt 21cmFAST}\footnote{https://github.com/andreimesinger/21cmFAST} \citep{Mesinger2007,Mesinger2011}. The simulations were performed on a cubic box of 128 comoving Mpc on each side, with $64^3$ grid cells. For this work, we use coeval boxes at redshift 12, and extract a single slice from each cube to form a 2D dataset.

\noindent \textbf{Network details.} To obtain hybrid summaries, we use a CNN with three convolutional layers (32, 64, and 128 filters) followed by max pooling. The flattened output is processed through two fully connected layers, with ReLU activations throughout. For classification, we use an FCN with hidden layers of sizes 128, 64, and 64, each followed by ReLU activation. For the EPE loss, a mixture density network (MDN) with a 64-unit hidden layer and output layers for mixing coefficients, standard deviations, and means is employed. Mixing coefficients use softmax, standard deviations are exponentiated, and means are directly outputted, with five components used.

\begin{figure}
  \centering
\includegraphics[width=0.84\linewidth]{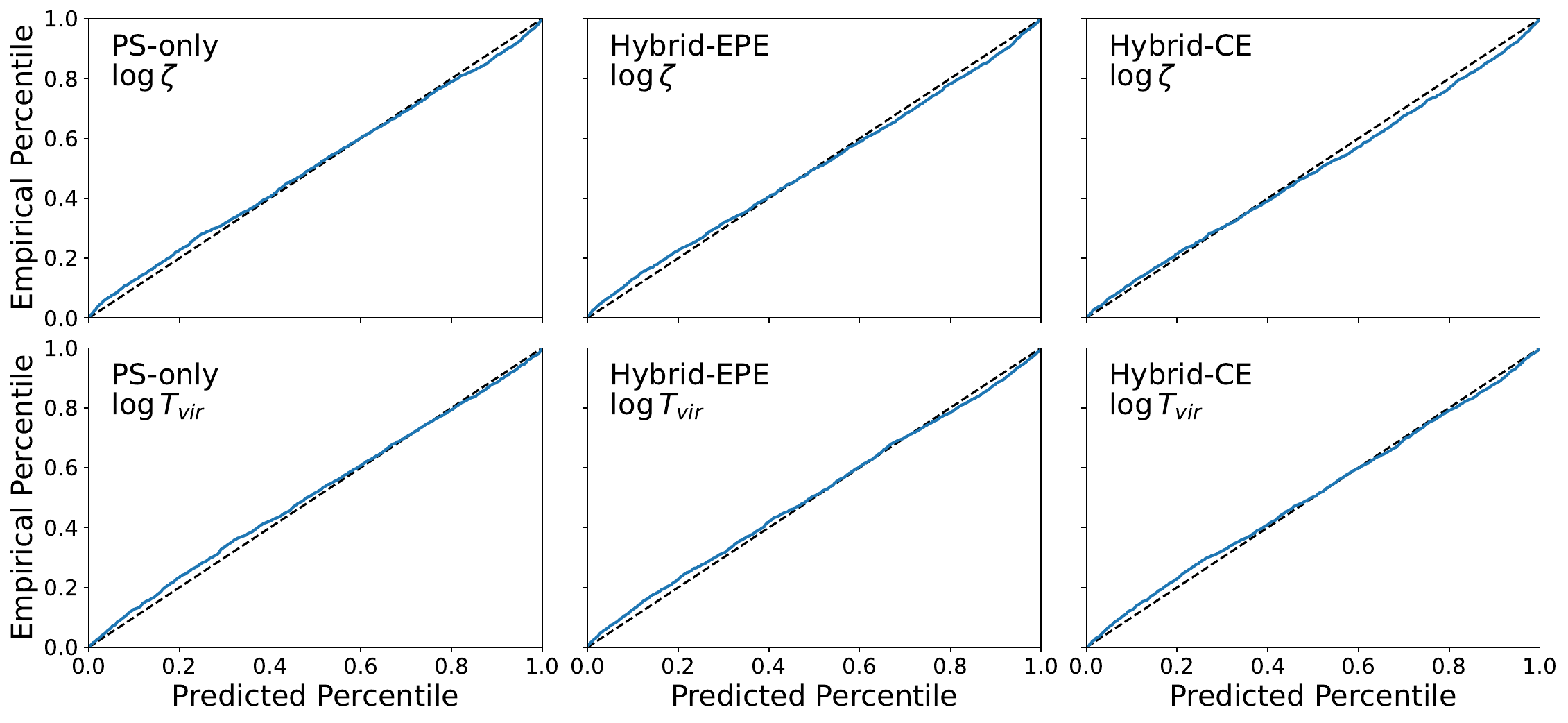}
  \caption{Posterior coverage for the 21 cm example across three summaries and two reionization parameters. The blue lines represent the actual calibration using 2,048 test samples, while the black dashed line indicates perfect calibration.}
  \label{fig:21cm_coverage}
\end{figure}

\noindent \textbf{Posterior Coverage.} In Figure~\ref{fig:21cm_results}, we present the posterior for a single test sample. To demonstrate that these results are not due to overfitting, we also show calibration results for a test set of 2,048 samples. Posterior coverage is used as a validation metric, as shown in Figure~\ref{fig:21cm_coverage}. The predicted percentiles closely match the empirical percentiles, indicating that our summaries are robust and the SBI inference is neither overly confident nor conservative in any case.

\subsection{Weak Lensing Simulations}
Here we test hybrid summaries on noisy tomographic WL simulations presented in \cite{makinen2024hybridsummarystatisticsneural}. The simulations were generated using the \texttt{pmwd} particle mesh code and then collected into four redshift bins to form convergence image data of shape $(128,128,4)$. The simulations are generated over a wide uniform prior in the parameters $p(\Omega_m, S_8) = \mathcal{U}([0.15, 0.7] \times [0.35, 1.52])$. As an existing summary, we repeat
\citet{makinen2024hybridsummarystatisticsneural}'s procedure and histogram all auto- and cross-power spectra for each redshift bin into a vector of 60 numbers for each simulation. The simulations are also subject to additive shape noise, which we add to the noise-free simulations on-the-fly during network training. 

\noindent \textbf{Network details.} For obtaining hybrid summaries the convolutional neural network with symmetric Multipole Kernels (MPK) was adapted from \citet{makinen2024hybridsummarystatisticsneural}. The lightweight network is initialised without pretraining for this analysis and contains $1,615$ learnable parameters. For the large non-hybrid CNN we apply a $3 \times 3$ kernel to embed the field into 16 filters, and then down-sample with stride-2 convolutions with output filters $[32, 64, 128]$. The network is then mean-pooled in the spatial axes and the flattened filters are passed to a dense network with a specified output size to be fed into the mutual information maximiser (here a mixture density network for the EPE loss). This results in $97,971$ learnable parameters. For both embedding networks we employ the \texttt{smooth\_leaky} activation function from \cite{makinen2024hybridsummarystatisticsneural}. For the EPE loss configuration, a mixture density network (MDN) with hidden layers of size $[70,70]$ and output layers for mixing coefficients, standard deviations, and means is employed to parameterise a four-component mixture.

\end{document}